\documentclass[sn-mathphys,Numbered]{sn-jnl}

\usepackage{graphicx}%
\usepackage{multirow}%
\usepackage{amsmath,amssymb,amsfonts}%
\usepackage{amsthm}%
\usepackage[title]{appendix}%
\usepackage[table]{xcolor}%
\usepackage{manyfoot}%
\usepackage{booktabs}%
\usepackage{listings}%

\usepackage{array}
\usepackage{longtable}
\usepackage{subcaption}%

\newcommand{\cline}{\cmidrule}

\begin{document}

\title[Article Title]{An experimental comparative study of backpropagation and alternatives for training binary neural networks for image classification}

\author*[1]{\fnm{Ben} \sur{Crulis}}\email{ben.crulis@univ-tours.fr}

\author[1,2]{\fnm{Barthelemy} \sur{Serres}}\email{barthelemy.serres@univ-tours.fr}
\equalcont{These authors contributed equally to this work.}

\author[1]{\fnm{Cyril} \spfx{de} \sur{Runz}}\email{cyril.derunz@univ-tours.fr}
\equalcont{These authors contributed equally to this work.}

\author[1]{\fnm{Gilles} \sur{Venturini}}\email{gilles.venturini@univ-tours.fr}
\equalcont{These authors contributed equally to this work.}

\affil*[1]{\orgdiv{LIFAT}, \orgname{University of Tours}, \orgaddress{\city{Tours}, \country{France}}}
\affil*[2]{\orgdiv{CETU ILIAD3}, \orgname{University of Tours}, \orgaddress{\city{Tours}, \country{France}}}

\abstract{Current artificial neural networks are trained with parameters encoded as floating point numbers that occupy lots of memory space at inference time. Due to the increase in the size of deep learning models, it is becoming very difficult to consider training and using artificial neural networks on edge devices.
Binary neural networks promise to reduce the size of deep neural network models, as well as to increase inference speed while decreasing energy consumption. Thus, they may allow the deployment of more powerful models on edge devices. However, binary neural networks are still proven to be difficult to train using the backpropagation-based gradient descent scheme.
This paper extends the work of \cite{crulis2023alternatives}, which proposed adapting to binary neural networks two promising alternatives to backpropagation originally designed for continuous neural networks, and experimented with them on simple image classification datasets. This paper proposes new experiments on the ImageNette dataset, compares three different model architectures for image classification, and adds two additional alternatives to backpropagation.
}


\maketitle

\section{Introduction}
Artificial Neural Networks (ANNs) are the main model architectures used to solve vision tasks nowadays. Unfortunately, they still require lots of computations to be trained and to perform inference. To achieve good task performance, they also tend to need lots of parameters, more than what is theoretically required. Their large number of parameters increases their size on disk and in memory, as well as decreases their inference speed. This makes ANN models difficult to deploy and use on edge devices such as smartphones.

Deploying ANN models on edge devices such as smartphones could lead to several benefits such as reducing the amount of possibly insecure communications with the cloud or relying on cloud computing in the first place. For this reason, the perspective of reducing the size of deep learning models is particularly appealing to store and use several larger high task performance vision models at the same time on resource-constrained devices.
If not only smaller models but also more efficient training methods are considered, the possibility of training ANNs directly on edge devices becomes conceivable. This would enable more possibilities in the domains of model personalization and user-centric deep learning while ensuring sensible personal data stays on the device at all times.

Binary Neural Networks (BNNs) consist in learning and deploying deep learning models with a much smaller memory footprint, faster inference, and reduced energy usage thanks to the use of efficient low-level binary operations. BNNs are thus very attractive for deployment in resource-constrained edge devices such as smartphones, while possibility benefiting from their eventual Graphical Processing Units\cite{He2021} or specialized hardware such as Field-Programmable Gate Array\cite{Zhou2017}.

BNNs have been used for object detection\cite{Zhao2022}, voice recognition\cite{Qian2019}, pedestrian detection \cite{ojeda2020}, stereo (depth) estimation\cite{Chen2020}, human activity recognition\cite{Daghero2021} and face mask wear and positioning correction\cite{Fasfous2021}. 

The main method used to train BNNs was introduced in \cite{Hubara2016a}. Their training scheme consist in slightly altering the forward and backward (backpropagation) passes of the regular training scheme to be able to get gradient flow through binarization steps, which would normally cause the gradient to be null. Due to the various approximations and heuristics used in this family of training schemes, training BNNs is still difficult and produces models with lower task performance compared to a model trained with full-precision weights.

However, recent advances in alternative training algorithms to backpropagation (BP) that match its task performance in full-precision networks lead us to consider using them for training BNNs as well.
To the best of our knowledge, trying alternatives to BP for training BNNs was not attempted before, except in \cite{crulis2023alternatives} of which this paper is an extension.
This previous paper experimented with two BP alternatives on smaller image classification tasks.

In this new work, we provide new experimental results on a harder image classification task and add two new alternative algorithms to the comparison. We also experiment with three different model architectures for image classification compared to the single feed-forward fully connected architecture previously considered.

We also examine the qualitative differences between the algorithms and architecture choices.

The main contributions of the paper are:
\begin{enumerate}
    \item We scale up the experiments of \cite{crulis2023alternatives} to a more realistic use case, adding new alternative algorithms and model architectures to the comparison.
    \item We provide extensive tests on the ImageNette dataset.
    \item We release an open source Pytorch framework that can be used to train binary and non-binary neural networks with different algorithms in a model agnostic manner\footnote{The code is available at \url{https://github.com/BenCrulis/binary_nn_extended}}.
\end{enumerate}


The rest of the paper is structured as follows.
Section \ref{background} presents several alternatives to the backpropagation algorithm that allow to train deep neural networks on supervised tasks. Then, Section \ref{algorithms} introduces the chosen method of binarization compatible with all training algorithms. Section \ref{experiments} details the experiments and results on the ImageNette dataset. Before summing up the paper in the conclusion, we discuss the results and their implications in Section \ref{discussion}.

\section{Background} \label{background}

In this section, we present the classical way of training Binary Neural Networks as well as some approaches that might replace the backpropagation mechanism.

\subsection{Binary Neural Networks}

This subsection introduces the main characteristics of BNNs. For more details about BNNs, please refer to recent surveys such as~\cite{yuan2023BNNSurvey}.

Contrary to classical ANNs with real-valued parameters, every single parameter of a binary neural network is coded on a single bit, which can be seen as an extreme form of quantization of the full-precision weights. A bit representing a single weight can encode one element among a pair of values, usually $-1$ and $1$ for binary neural networks as it allows to replace the floating point multiplication by a logical XNOR operation.
The immediate consequence is that binary neural networks are much smaller than their continuous counterparts, one can theoretically expect a reduction in the size of a factor $32$ compared to a network using $32$ bits floating point weights \cite{yuan2023BNNSurvey}.
Provided the activation function used in the hidden layers also outputs binary values, the following computation of the next binary layer can be optimized with efficient binary operation and also provide inference speedups.

\paragraph{Training binary neural networks using gradient descent}

Training a multi-layer BNN by searching naively for a valid set of binary weights is a combinatorial problem that does not scale due to its non-linear binary optimization nature ~\cite{BLUM1992117}.

For this reason, to train deep BNNs, \cite{Hubara2016a} proposes to train a model with full-precision weights as a support for the binarized model.

The classical BP algorithm is modified to allow computing a non-zero gradient of the loss with respect to the parameters when traversing binary activation layers using the Straight-Through Estimator (STE). This latter turns the unusable derivative of the $sign$ function into a useful approximation, for instance, an identity function.

Then, the deep neural network training is done following the classic manner with few modifications~\cite{Hubara2016a}.

The forward pass binarizes the continuous model parameters. For that, in our case, we use the $sign$ function that outputs $-1$ or $1$.
These binary weights are used to compute the predicted values and the loss. During this forward pass, the data can also pass through binary activation layers (e.g. the $sign$ function).

The binarization steps of the backward pass use the chosen STE to compute the gradient that will be used to update the real-valued model parameters.

As in the classical backpropagation scheme, the gradient of the input with respect to the output is computed for each layer using the binary weights to send the gradient information upstream.

Please note that the gradient computed in the backward pass is generally not binary.
When deploying the model, the full-precision parameters can be used to compute the final binary weights using the $sign$ function and discarded, reducing the model size for both storage and inference.

The activation function of a BNN can be any differentiable function 
(RELU, GeLU, etc.), or any non-differentiable function (step functions, for instance) 
if an appropriate estimator of the gradient is provided.
However, only the activation functions outputting binary vectors, e.g. $sign$ function, lead to storage reductions for intermediate activation values. Intermediate binary activation values also allow obtaining inference speed gains if the following layer is a binary linear or convolutional layer, thanks to the possibility of rewriting the floating point multiply and accumulation operations into efficient low-level binary operations, namely XNOR and bitcount.
The former point is particularly important in convolutional neural networks since the activation buffers usually take much more space than the kernels used to compute them, so a reduction in the activation storage requirements can be very beneficial at inference time.

The difficulty of training BNNs comes from the approximations that are made to enable training of the parameters using gradients. The STE being one of these approximations used in the binary activation layers, in a normal backpropagation pass, the error signal will flow recursively through several of these estimators. We conjecture that having the gradient traverse several of these estimators decreases the quality of the teaching signal to the point it starts to hurt the learning performance of the model.

\subsection{Backpropagation and alternatives}

\paragraph{Backpropagation}

BP is the most common method for the end-to-end training of ANNs. BP allows to decrease the error of a differentiable model by performing Gradient Descent (GD) on the loss landscape induced by the differentiable training loss objective. After a forward pass on a data batch and the computation of the loss, a gradient of the parameters with respect to the loss is computed in all layers using the chain rule, in a step called the backward pass. This gradient is then used to apply a small modification of the weights which slightly improves the model performance on the training set. This training scheme is illustrated in Fig. \ref{fig:algorithms}a.

\paragraph{Feedback Alignment}
Feedback Alignment (FA) is an alternative to backpropagation introduced in \cite{Lillicrap2016} with the explicit goal of proposing a more biologically plausible learning mechanism for training neural networks. FA demonstrates the surprising fact that, under some conditions, backpropagating the error of the output layer using the transpose matrix of each linear layer is not needed, and that a constant random matrix is sufficient to provide a useful learning signal to the upstream layers.
By using random matrices, one might have the intuition that no useful features could be learned in the upstream layers. Yet, \cite{Lillicrap2016}~shows that the weight matrices adapt to the constant random matrices and, in turn, allow the network to extract useful features in all layers. 

\paragraph{Direct Feedback Alignment}
In \cite{Nokland2016}, the author goes a step further and proposes Direct Feedback Alignment (DFA) a method that improves on FA by propagating the output error signal directly to each layer using constant random matrices. Contrary to FA where the error signal goes through up to $K-1$ layers, where $K$ is the number of layers in the network, in DFA the output error goes through exactly $0$ layer in reverse.
As the error effectively skips all downstream layers to train a particular layer, this method allows the training of very deep networks (with more than $100$ layers) where BP would fail to converge. Although \cite{Nokland2016} suggests this is due to the difference in the paramter initialization, it is also explained if downstream layers are frozen, DFA cannot decrease the error whereas BP can.
This training scheme is illustrated in Fig. \ref{fig:algorithms}b.

\paragraph{Direct Random Target Projection}
All previous methods necessitated computing the output loss of the model in order to provide an error signal to train the layers. This effectively prevents the model's parameters from being updated until the forward pass is completed, this phenomenon is referred to as \textit{update locking}. However, another algorithm called Direct Random Target Projection (DRTP) was recently introduced to address this problem \cite{Frenkel2021} and provide a less costly and more biologically plausible learning algorithm for deep neural networks. DRTP projects the target labels directly to each layer using constant random matrices in a way similar to DFA and uses the result as a learning signal to train each layer. By design, DRTP allows each layer to be updated as soon as its input activations are available, it is said to be \textit{update-unlocked}. The parameters of models trained using DRTP can be updated while the forward pass is not completed, which also saves memory as the input buffers can be released once the layer parameters are updated. 
This training scheme is illustrated in Fig. \ref{fig:algorithms}c.

\paragraph{HSIC Bottleneck}
The HSIC Bottleneck method \cite{KurtMa2020, Pogodin2020} uses the Hilbert Schmidt Independence Criterion (HSIC) to create a kernelized layer-wise loss function that provides a signal for learning each layer in parallel. The loss function encourages each layer to compress the information coming from the input while increasing the dependence of the layer output with the target labels. The balance between these two terms is controlled using the $\gamma$ parameter. As for DRTP, this method is update-unlocked.
This training scheme is illustrated in Fig. \ref{fig:algorithms}d.

\paragraph{SigpropTL}
The Sigprop framework \cite{Kohan2022} corresponds to a family of alternative algorithms that propagate the learning signal in a forward pass in the same manner as the input data used for the inference, thus bypassing the need for a backward pass. We focus on the Target Loop variant of the algorithm that computes an error at the output layer and sends it back to the input layer using a constant random projection matrix. Each layer is then updated to make the output closer to the computed target. This method is very similar to the Indirect Feedback Alignment method described in \cite{Nokland2016}. If the input is passed again in the second forward pass along with the error signal, the method can be implemented in an update-unlocked manner as saving input buffers becomes unnecessary.
This training scheme is illustrated in Fig. \ref{fig:algorithms}e.

\paragraph{Choice of algorithms} These algorithms (DFA, DTRP, HSIC, SigpropTL) were chosen for the experiments because they fulfill the following criterions:

\begin{enumerate}
    \item They allow the training of deep neural network models of any feed forward neural network architecture.
    \item They can be made compatible with binary neural networks.
    \item They are supervised learning algorithms.
    \item They all compute teaching signals in a way that avoid recursively computing the derivatives of the activation functions.
    \item They all have been showed to have reduced computational or memory cost compared to BP.
\end{enumerate}

\paragraph{Other candidate alternatives for training BNNs}
Very different types of training algorithms have been proposed to overcome the limitations of BP or to provide biologically plausible alternatives. %

Some approaches try to view synapses or neurons as Reinforcement Learning agents that learn to predict the weight modifications that will improve the loss~\cite{Lansdell2020}.
Other methods manage to get notable results by performing random modifications of the model parameters and accepting or rejecting the modifications, e.g. \cite{Akshat2022}.

Due to their greater computational costs and difficulty in reconciling with BNNs, we leave these other algorithms out of our experiments.

\section{Algorithms} \label{algorithms}

\begin{figure*}[t]
\centerline{\includegraphics[width=1.0\textwidth]{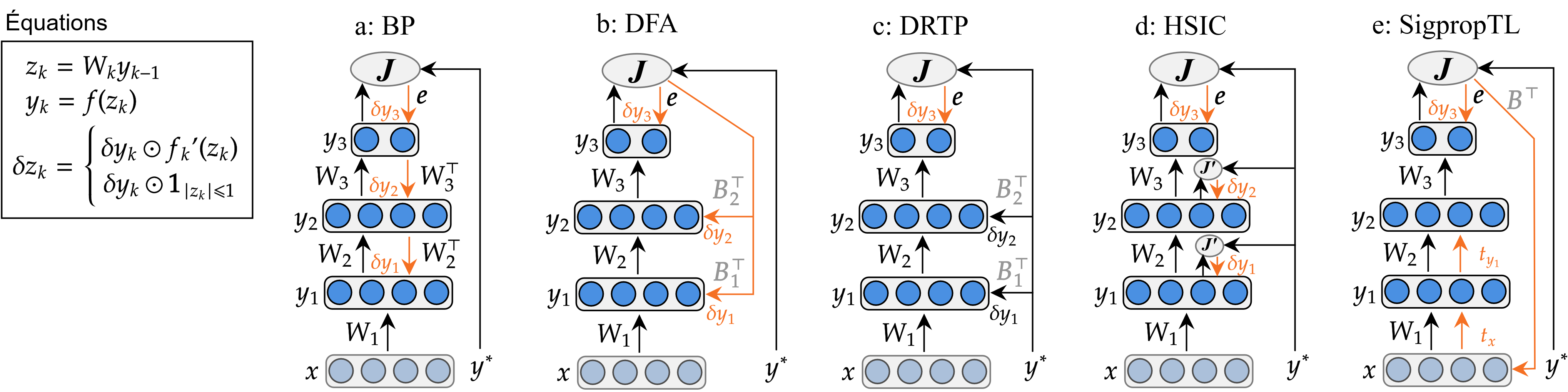}}
\caption{Summary of the algorithms tested in the following experiments. The $B_k$ matrices are constant random matrices 
initialized once before starting to train the model. Adapted from \cite{Frenkel2021}.}
\label{fig:algorithms}
\end{figure*}

Figure \ref{fig:algorithms} summarizes the differences between the chosen algorithms. BP is the only algorithm tested where the error traverses the layers in reverse, the other algorithms directly send an error to each layer in parallel. BP (a), DFA (b), and SigpropTL (d) are the only algorithms where the information learned in the downstream layers affects the computation of the errors in the upstream layers, although only indirectly in the case of DFA and SigpropTL. DRTP (c) and HSIC (d) can directly train the layers in the first forward pass, the other algorithms all require at least two passes.

\subsection{Binarization of the weights and activations}

\begin{figure*}[t]
\centerline{\includegraphics[width=0.6\textwidth]{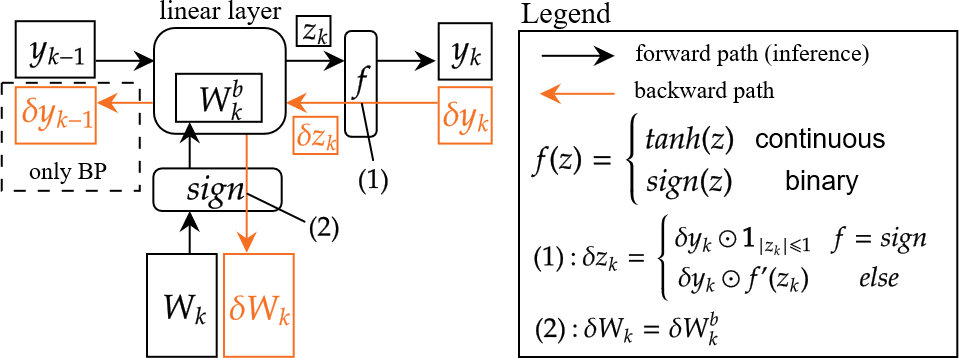}}
\caption{Summary of the chosen binarization scheme effective at each binary weight layer. Here a single layer of the model at index $k$ is represented with both the linear layer using the binary weights $W^b_k$ and the activation function $f$.}
\label{binarization}
\end{figure*}

In order to make the algorithms compatible with weight binarization, we slightly modify the way the forward pass and backward pass are computed.
Firstly, we change the activation function of the hidden layers to be a $sign$ function with a Straight-Through Estimator (STE), i.e. the derivative of the $sign$ function is replaced with another function that will provide a non-zero gradient to previous layers.
In the experiments, we use two variants of the STE. Let $\delta y_k$ be the incoming gradient at layer $k$ and $\delta z_k$ the estimated output gradient of the $sign$ function. One of the STE variants is ignoring the gradient of the $sign$ function as if the activation was the identity function, so in this case $\delta z_k = \delta y_k$. This is the non-saturating STE. Another variant introduced in \cite{Hubara2016a} is equivalent to propagating the gradient through a hard $tanh$ function ($Htanh(x) = clip(x, -1, 1)$), in which case $\delta z_k = \delta y_k \odot \mathbf{1}_{\vert z_k \vert \le 1}$. This STE is saturating as the gradient is $0$ when the neuron has reached saturation, that is when $\vert x \vert > 1$, $x$ being the neuron pre-activation.

In the experiments, we use the non-saturating STE for the binarization of the weights when applicable and the saturating STE for the activation function when using the $sign$ function as the activation function of the hidden layers.

Figure \ref{binarization} shows the flow of information in the forward and backward pass when learning. The vector $z_k$ is the pre-activation and $y_k$ is the activation at layer $k$. When in the forward pass, the real valued weights are first binarized to get the binary weight matrix $W^b_k$ using the $sign$ function, and then used to compute the pre-activations vector $z_k$ from $y_{k-1}$ in order to get $y_k$ as follows: $y_k=f(W^b_k y_{k-1})$. This value is then used as input to the next layer or as the network output.

In the backward pass, the error $\delta y_k$ is sent through the STE of the activation function if it is binary, or computed normally if the activation function is $tanh$. This gives us the error $\delta z_k$ that is itself used together with $y_{k-1}$ to compute the gradient of the real-valued weights $\delta W_k$ by sending it through the non-saturating STE.
In the particular case of the BP algorithm, the error $\delta z_k$ is also used to compute the output error of the previous layer of index $k-1$ as follows: $\delta y_{k-1} = (W^b_k)^\top \delta z_k$. This means the error signal that already went through an STE will go through another one upon reaching the previous layer. The parameters at layer $k$ thus receive an error that went through $K-k$ STEs, $K$ being the total number of layers in the model. The other algorithms all compute the error $\delta y_k$ either from the loss error or directly from the target labels $y^*$ and so receive an error that went through a constant number of STEs, only once in our case.

\section{Experiments} \label{experiments}

In this section, we present a test protocol to compare the training of binary neural networks with the different algorithms.

We propose to explore the following questions:
\begin{enumerate}
    \item How much does performance degrade when we binarize reference vision classifier models as measured in term of accuracy?
    \item Which between binarizing the weights and the activations degrades the task performance the most?
    \item Can we improve the BNN task performance using BP alternatives for some architectures?
    \item How skip-connections do impact the BNN training of models using BP and its alternatives?
    \item What are the runtime and memory costs of BP alternatives for training BNNs?
\end{enumerate}

\subsection{Model architectures}

We compare $3$ model architectures for image classification, VGG-$19$, MLP-Mixer and MobileNetV$2$. The biggest model is the VGG-$19$ model~\cite{Simonyan2015} and the smallest is MobileNetV$2$~\cite{Sandler2018} which introduces inverted residual layers and bottleneck layers to reduce the size of the model. Both of these models are convolutional neural networks (CNNs) which uses learnable convolutional filters to process image data.

As some alternatives to the backpropagation such as DFA were found to have trouble optimizing CNNs, we add a third non-convolutional model called MLP-Mixer~\cite{Tolstikhin2021} to the experiments. MLP-Mixer uses only dense layers and transpose operations to perform feature extraction on image patches while allowing information to mix between spatial locations and thus extracts higher level features useful for the classification.

For the different algorithms that require a notion of layer output to have a point to send signals to, we assume the output of a layer correspond to the last operation being executed before reaching either the next fully connected or convolutional layer in the computational graph of the model. In general, the operation satisfying this criterion will be an activation function. The teaching signals are sent back from these locations and stop propagating when reaching the output of the previous layer. This method allows our framework to work with less regular architectures and stay model-agnostic.

Another point worth noting is that both MobileNetV2 and MLP-Mixer use skip connections\footnote{also know as residual connections} which can prevent the use of optimized XNOR operations in some of the layers. Indeed, whether the connection is made from a binary output or not, the sum operation will necessarily be a floating point operation or an integer operation at best, preventing the use of low level binary operations. 

\subsection{Protocol}

We now propose to measure the relative difference in terms of task performance of the algorithms for training binary neural networks on the ImageNette dataset, the Indoor Recognition Dataset and CIFAR-$10$.
The ImageNette dataset is a $10$-class subset of ImageNet with $9488$ training images and $3944$ test images.
The Indoor Scene Recognition Dataset contains $67$ scene classes with $14280$ training images and $1340$ test images. The CIFAR-$10$ dataset contains $10$ classes with $50000$ training images and $10000$ test images.
In the remainder, the training set is divided with a $90\%/10\%$ split for training and validation.

The test framework is implemented in PyTorch and makes heavy use of the injection hooks in forward and backward passes to implement the different 
algorithms in a model-agnostic manner. The loss function used is a standard cross entropy loss and a batch size of $128$ is used in all experiments. Due to 
Pytorch limitations regarding the use of the low-level binary operations and binary data types, we implement the BNNs using floating point numbers.

When binarizing layers, all activation functions are replaced by the $sign$ function, the linear and convolutional layer weights are binarized, with the exception of biases. The first and last layers of the models are not binarized. The training images are augmented using random horizontal flips, random rotation offsets and random crops.

For each algorithm and binarization configuration, we first run a grid search on the learning rate and algorithm-specific hyperparameters to find the combinations leading to the highest accuracy on the validation set in $80$ epochs. The possible values for the learning rate $\mu$ are $10^{-3}$, $10^{-4}$ and $10^{-5}$.
For HSIC, the possible values for $\gamma$ are $2$, $20$, and $200$. For Sigprop, we search for a scalar factor $\alpha$ for the scale of the random matrix among the values of $1$, $0.1$ and $0.01$.

A summary of the algorithm configuration is given in Table \ref{configuration} and a summary of the models can be found in Table \ref{model_summary}.
For each configuration, we re-run the experiments $5$ times and report means and standard deviations for the accuracy on the test set.

\begin{table}
\centering
\caption{Configuration for the experiments}
\label{configuration}
\begin{tabular}{r|lllll} \toprule
\multicolumn{1}{l}{}        & BP                        & DFA                       & DRTP                      & HSIC                              & SigpropTL                          \\ \hline
grid search hyperparameters & \multicolumn{1}{c}{$\mu$} & \multicolumn{1}{c}{$\mu$} & \multicolumn{1}{c}{$\mu$} & \multicolumn{1}{c}{$\mu, \gamma$} & \multicolumn{1}{c}{$\mu, \alpha$}  \\
weight decay                & \multicolumn{5}{c}{$10^{-5}$}                                                                                                                              \\
optimizer                   & \multicolumn{5}{c}{Adam}                                                                                                                                   \\
Adam $\beta_1, \beta_2$     & \multicolumn{5}{c}{$0.9, 0.999$}                                                                                                                           \\ \bottomrule
\end{tabular}
\end{table}

\begin{table}
\centering
\caption{Summary of the models used in the experiments}
\label{model_summary}
\begin{tabular}{r|ccc} \toprule
\multicolumn{1}{l}{}        & MobileNet-V2  & MLP-Mixer   & VGG-19         \\ \hline
number of parameters         & $2,236,682$   & $7,279,098$ & $139,622,218$  \\
architecture                 & convolutional & dense       & convolutional  \\
has skip connections         & yes           & yes         & no             \\
original activation function & RELU          & GELU        & RELU           \\
normalization                & batchnorm     & layernorm   & batchnorm      \\ \bottomrule
\end{tabular}
\end{table}

The BP algorithm is implemented using regular Pytorch code while the other algorithms are optimized for memory usage, but not for parallelism. This is particularly important for DFA which is not expected to have reduced memory usage but only a potential for better parallelization of the backward pass.

\subsection{Results}

In this section, we try to reproduce the results from \cite{crulis2023alternatives} at a larger scale. That is, we compare the accuracy performance of the continuous models with their binarized versions.
We first conduct experiments on ImageNette and the Indoor Scene Recognition dataset to assess the performance degradation of BNNs with the main three algorithms: BP, DFA and DRTP.
Then, we study the impact of removing skip-connections on fully binarized models.
Finally, we add two new algorithms, HSIC and SigpropTL to the experiments and measure the run times and memory costs while training.

\paragraph{Binary models have degraded performance compared to continuous models}

We report the final performance of models trained with continuous weights and continuous activations along models with binary weights and binary activations.

As reported in Table \ref{result_summary}, BNNs have vastly inferior performance compared to their continuous counterparts in both datasets. The DRTP algorithm seems less affected by the binarization of the models to the point that its performance sometimes even improves, but it is the weakest overall.
We note that DFA manages to train better binary VGG-$19$ models compared to BP for all datasets.

\begin{table}
\centering
\caption{Accuracy scores (\%) on five independent runs. Standard deviations between parenthesis.}
\label{result_summary}
\begin{tabular}{r|cccccc} \toprule
\multicolumn{7}{c}{ImageNette}                                                                                        \\
\multicolumn{1}{l}{}             & \multicolumn{2}{c}{BP}    & \multicolumn{2}{c}{DFA}   & \multicolumn{2}{c}{DRTP}   \\
\multicolumn{1}{r}{Model} & Continuous       & Binary        & Continuous       & Binary        & Continuous       & Binary         \\ \hline
MLP-Mixer                        & 73.9 (0.81) & 61.7 (0.78) & 46.4 (1.32) & 36.6 (2.57) & 24.0 (2.67) & 26.0 (1.63)  \\
MobileNet-V2                     & 87.9 (0.34) & 80.4 (0.34) & 62.4 (0.50) & 52.9 (1.82) & 38.8 (0.47) & 29.6 (0.44)  \\
VGG-19                           & 89.1 (0.46) & 23.6 (0.95) & 69.0 (0.27) & 67.2 (0.67) & 43.2 (0.81) & 41.9 (1.16)  \\ \hline
\multicolumn{7}{c}{Indoor Scene Recognition}                                                                          \\
\multicolumn{1}{r}{}             & \multicolumn{2}{c}{BP}    & \multicolumn{2}{c}{DFA}   & \multicolumn{2}{c}{DRTP}   \\
\multicolumn{1}{r}{Model} & Continuous       & Binary        & Continuous       & Binary        & Continuous       & Binary         \\ \hline
MLP-Mixer                        & 32.5 (0.84) & 10.0 (2.01) & 10.2 (0.41) & 6.7 (0.43)  & 3.0 (0.40)  & 3.7 (0.46)   \\
MobileNet-V2                     & 58.4 (0.48) & 15.1 (0.98) & 21.3 (0.58) & 10.0 (0.37) & 7.7 (0.29)  & 4.5 (0.19)   \\
VGG-19                           & 58.7 (0.59) & 2.0 (0.14)  & 5.8 (0.52)  & 7.6 (0.27)  & 3.3 (0.25)  & 5.6 (0.09)   \\ \hline
\multicolumn{7}{c}{CIFAR-10}                                                                                          \\
\multicolumn{1}{c}{}             & \multicolumn{2}{c}{BP}    & \multicolumn{2}{c}{DFA}   & \multicolumn{2}{c}{DRTP}   \\
\multicolumn{1}{r}{Model} & Continuous       & Binary        & Continuous       & Binary        & Continuous       & Binary         \\ \hline
MLP-Mixer                        & 66.8 (0.59) & 48.3 (0.26) & 49.4 (1.47) & 46.1 (1.53) & 25.7 (1.71) & 37.9 (0.68)  \\
MobileNet-V2                     & 87.3 (0.27) & 58.0 (1.41) & 59.1 (0.90) & 45.5 (0.79) & 28.6 (0.92) & 28.0 (0.99)  \\
VGG-19                           & 89.5 (0.22) & 20.8 (1.76) & 44.8 (1.16) & 49.5 (0.69) & 27.7 (1.17) & 37.1 (0.61)  \\ \bottomrule
\end{tabular}
\end{table}

\paragraph{Breakdown of the effect of binarizing weights and activation separately}

We focus on the results on the ImageNette dataset and explore the effect of binarizing weights and activations independently to find the greatest degradation factor. We highlight in bold the algorithm that performs best for a given combination of weights and activations.
MobileNet-V$2$ and MLP-Mixer are better trained by the BP algorithm in all cases, as reported in tables \ref{mobilenetv2_results} and   \ref{mlpmixer_results} respectively. However, for VGG-$19$, DFA train models of much higher performance when weights are binary, regardless of the binarization of activations, as shown in Table \ref{vgg19_results}. No model with either binary weights or activation manages to reach the performance of models with continuous weights and activations, no matter the training algorithm.

\begin{table}
\centering
\caption{Accuracy scores (\%) for MobileNet-V2 on five independent runs on ImageNette. Standard deviations between parenthesis.}
\label{mobilenetv2_results}
\begin{tabular}{rr|lll} \toprule
weights                 & activations & \multicolumn{1}{c}{BP} & \multicolumn{1}{c}{DFA} & \multicolumn{1}{c}{DRTP}  \\ \cline{1-1}\cmidrule{2-5}
\multirow{2}{*}{binary} & binary      & \textbf{57.5} (0.66)   & 45.7 (1.41)             & 31.4 (1.13)               \\
                        & f32         & \textbf{80.4} (0.34)   & 52.9 (1.82)             & 29.6 (0.44)               \\
\multirow{2}{*}{f32}    & binary      & \textbf{64.4} (0.57)   & 53.7 (1.06)             & 39.5 (1.58)               \\
                        & f32         & \textbf{87.9} (0.34)   & 62.4 (0.50)             & 38.8 (0.47)               \\ \bottomrule
\end{tabular}
\end{table}

\begin{table}
\centering
\caption{Accuracy scores (\%) for MLP-Mixer on five independent runs on ImageNette. Standard deviations between parenthesis.}
\label{mlpmixer_results}
\begin{tabular}{rr|lll} \toprule
weights                 & activations & \multicolumn{1}{c}{BP} & \multicolumn{1}{c}{DFA} & \multicolumn{1}{c}{DRTP}  \\ \midrule
\multirow{2}{*}{binary} & binary      & \textbf{40.5} (0.77)   & 37.2 (1.58)             & 37.4 (0.53)               \\
                        & f32         & \textbf{61.7} (0.78)   & 36.6 (2.57)             & 26.0 (1.63)               \\
\multirow{2}{*}{f32}    & binary      & \textbf{65.7} (0.69)   & 47.5 (0.25)             & 36.9 (1.06)               \\
                        & f32         & \textbf{73.9} (0.81)   & 46.4 (1.32)             & 24.0 (2.67)               \\ \bottomrule
\end{tabular}
\end{table}

\begin{table}
\centering
\caption{Accuracy scores (\%) for VGG-19 on five independent runs on ImageNette. Standard deviations between parenthesis.}
\label{vgg19_results}
\begin{tabular}{rr|lll} \toprule
weights                 & activations & \multicolumn{1}{c}{BP} & \multicolumn{1}{c}{DFA} & \multicolumn{1}{c}{DRTP}  \\ \midrule
\multirow{2}{*}{binary} & binary      & 21.2 (0.53)            & \textbf{52.0} (0.79)    & 41.2 (0.54)               \\
                        & f32         & 23.6 (0.95)            & \textbf{67.2} (0.67)    & 41.9 (1.16)               \\
\multirow{2}{*}{f32}    & binary      & \textbf{65.6} (0.76)   & 61.8 (0.94)             & 48.8 (0.28)               \\
                        & f32         & \textbf{89.1} (0.46)   & 69.0 (0.27)             & 43.2 (0.81)               \\ \bottomrule
\end{tabular}
\end{table}

\paragraph{Performance of BNNs with all algorithms}

Table \ref{binary_results_ImageNette} centralizes all of the results of binary models on ImageNette. The values in bold indicate the best algorithm for the model. In the case of the binary VGG-$19$, BP and SigpropTL both fail to train compared to the other algorithms. However, since BP is the only algorithm of the two that reaches satisfactory performance with the other models, it suggests us to find a difference between VGG-$19$ and the other models instead.

\begin{table}
\centering
\caption{Accuracy scores (\%) for experiments with both binary weights and binary activations on ImageNette. Standard deviations between parenthesis.}
\label{binary_results_ImageNette}
\begin{tabular}{r|ccccc} \toprule
\multicolumn{1}{l}{} & BP                   & DFA                  & DRTP        & HSIC        & SigpropTL    \\ \hline
MLP-Mixer            & \textbf{40.5} (0.77) & 37.2 (1.58)          & 37.4 (0.53) & 25.0 (0.59) & 20.9 (1.17)  \\
MobileNet-V2         & \textbf{57.5} (0.66) & 45.7 (1.41)          & 31.4 (1.13) & 25.9 (0.58) & 12.0 (0.75)  \\
VGG-19               & 21.2 (0.53)          & \textbf{52.0} (0.79) & 41.2 (0.54) & 28.0 (0.17) & 14.1 (2.05)  \\ \bottomrule
\end{tabular}
\end{table}

\paragraph{Skip-connections have a large impact on the performance of binary models}

Compared to Mobilenet-V2 and MLP-Mixer, VGG-$19$ does not have any skip-connection that makes it easier for information to flow forward and backward in the model.
We thus try to see what happens to the performance of the MLP-Mixer and Mobilenet-V2 models when removing the skip connections used in their hidden layers and report the results in Table \ref{binary_no_skip_results}. In this case, DFA trains the better-performing models even though BP achieves better performance for these two models when using the skip connections.

\begin{table}
\centering
\caption{Accuracy scores (\%) of BNN models after removing skip-connections on five independent runs on ImageNette. The VGG-$19$ results are reported from Table \ref{binary_results_ImageNette} as VGG-$19$ already has no skip-connections. Standard deviations between parenthesis.}
\label{binary_no_skip_results}
\begin{tabular}{r|ccccc} \toprule
\multicolumn{1}{l}{} & BP          & DFA                  & DRTP        & HSIC        & SigpropTL    \\ \hline
MLP-Mixer            & 18.2 (0.37) & \textbf{26.5} (3.39) & 22.0 (1.42) & 20.7 (1.04) & 13.8 (1.24)  \\
MobileNet-V2         & 34.4 (1.15) & \textbf{40.6} (1.22) & 32.1 (1.07) & 25.2 (0.32) & 11.0 (0.24)  \\
\textit{VGG-19}               & \textit{21.2 (0.53)}          & \textit{\textbf{52.0} (0.79)} & \textit{41.2 (0.54)} & \textit{28.0 (0.17)} & \textit{14.1 (2.05)}  \\ \bottomrule
\end{tabular}
\end{table}

\paragraph{Run time performance and memory costs}

Table \ref{system_statistics} shows the peak memory usage and run times for the different algorithms while training on ImageNette.
As the experiments were run on two different machines, we reported the results by a graphics card (A$6000$ and Quadro RTX $8000$ respectively). 
Regarding memory, BP and DFA appear to have similar costs while the other $3$ have much lower costs. DRTP and HSIC achieve the lowest memory costs depending on the type of graphics card used. The lowest training cost is achieved while training MLP-Mixer using HSIC, requiring only $2.1$GB of memory.

Regarding the run times, once again BP and DFA are comparable and DRTP also achieves similar figures. However, the fastest algorithms appear to be HSIC and SigpropTL.

\begin{table}
\centering
\caption{Peak memory usage and total run times on ImageNette. Standard deviations between parenthesis.}
\label{system_statistics}
\setlength{\tabcolsep}{0.47mm}
\begin{tabular}{rr|ccccc}
\toprule
                                 &                      & BP           & DFA          & DRTP               & HSIC                 & SigpropTL           \\ \hline
\multirow{3}{*}{GPU memory (GB)} & MLP-Mixer            & 6.7 (0.0)    & 6.7 (0.0)    & 2.8 (0.0)          & \textbf{2.1} (0.0)   & 2.7 (0.0)           \\
                                 & MobileNet-V2         & 7.3 (0.0)    & 7.3 (0.0)    & 4.5 (0.0)          & \textbf{4.4} (0.0)   & 5.5 (0.0)           \\
                                 & VGG19                & 21.1 (0.0)   & 21.1 (0.0)   & \textbf{9.4} (0.0) & 10.8 (0.0)           & 12.4 (0.0)          \\ \hline
\multirow{3}{*}{Run time (s)}    & MLP-Mixer            & 12905 (502)  & 13772 (1170) & 13577 (54)         & 9341 (57)            & \textbf{8270 }(91)  \\
                                 & MobileNet-V2         & 12995 (1105) & 12855 (240)  & 13836 (483)        & 9014 (116)           & \textbf{8095 }(95)  \\
                                 & VGG19                & 17057 (836)  & 17915 (75)   & 20301 (3282)       & \textbf{12087 }(168) & 12830 (94)          \\ \bottomrule
\end{tabular}

\end{table}

\section{Discussion} \label{discussion}

\paragraph{What is the impact of binarizing weights and activations on the accuracy?}
Binarizing the activations and binarizing the weights appear to have a roughly similar negative impact on performance when using BP, except when considering the VGG-19 model which appears to be much more affected by the binarization of the weights. For DFA and DRTP, the binarization of weights and activations has a much smaller impact. These results suggest that it might be advantageous to compromise by only binarizing either the weights or the activations to decrease the accuracy loss while saving memory at inference time.

\paragraph{How do the algorithms compare when training BNNs?}
For the MLP-Mixer and MobileNet-V2 models, BP trains the most accurate models, although it still reaches a much lower accuracy compared to the best model that is non-binary. For these two models, DFA and DRTP achieve similar performance, close to BP in the case of MLP-Mixer. HSIC and SigpropTL produce the worst-performing models.
BP fails to get the best performance for the VGG-19 model, while DFA achieves accuracy $30$ points greater. BP is even outperformed by DRTP and HSIC in this case.
This suggests VGG-$19$ has a particular architecture that makes it hard for BP to optimize the BNN model but does not affect the other algorithms as much. This might be due to exploding or vanishing gradient issues due to the large size of convolutional kernels of this model compared to MobileNet-V2 and the dense layers of MLP-Mixer.

\paragraph{Is having skip-connections important to reach good performance with BNNs?}
Removing the skip connections highly degrades the performance of all the algorithms. However, BP shows the biggest performance drop, to the point it makes DFA the better performing algorithm with these architecture variants for both the MLP-Mixer and MobileNet-V2 models.

Surprisingly, without skip-connections the best results are reached by the VGG-$19$ model with DFA ($52.0\%$ accuracy), although DFA was previously observed to not work well with convolutional neural networks in the literature.

\paragraph{Which method(s) should be preferred and which ones should be avoided for training BNN models}

BP appears to obtain the best performing BNN models when using the more modern neural network architectures that have features like skip-connections. However, BP falls under DFA when we consider the older VGG-19 model or when removing skip-connections. Due to the update locking of the weights, BP and DFA have the highest memory costs, while the other $3$ update unlocked algorithms reduce the memory usage by a factor of more than $2$ in nearly all cases. DRTP, HSIC and SigpropTL are also the fastest training algorithm, although HSIC and SigpropTL failed to achieve competitive performance in term of accuracy.

Overall, if accuracy is the sole goal for the BNN training, BP should probably still be the preferred algorithm, even though it might be worthwhile to also attempt training with DFA in case the particular architecture considered is more compatible with it than with BP. If the memory cost or training run times are more important, then DRTP should be the preferred candidate algorithm, although the resulting accuracy of the model is greatly negatively impacted by this algorithm. It should be noted that all these algorithms have run for the same number of epochs, and thus it is possible that higher performance could be reached with some of the algorithms by running them longer to ensure convergence.

\section{Conclusion}

Binary Neural Networks are more suited to use on edge devices compared to full-precision networks because of their lower memory and computational costs as well as their lower energy usage. 
The current approach to deal with BNN learning is to use backpropagation with the STE. Nevertheless, several alternatives to backpropagation, such as DFA and DRTP, were recently proposed but not tested in the context of BNNs. These alternatives have lower complexity and memory cost in comparison to backpropagation. To the best of our knowledge, we are the first to study the impact of the different learning schemes and their binarization in terms of performance (accuracy) on a medium sized image classification tasks. 

Contrary to what was previously found with simpler models on the image datasets MNIST, FashionMNIST and CIFAR-$10$, it appears that BP still mostly outperforms the other algorithms when training BNNs for the more modern architecture variants on ImageNette. However, if we consider training older architectures or if variants of the architectures are considered, it is possible that BP fails and that we can obtain better accuracy using DFA or DRTP.
If accuracy is a secondary objective, the alternative algorithms have some computational advantages by design, opening up possibility of training binary neural networks directly on edge devices. In the future we will exploit these ideas to train and deploy more efficient models on edge devices, in particular, vision models on smartphones.

\section*{Compliance with Ethical Standards}
The authors declare that they have no conflict of interest. The research was not funded. The experiments did not involve human or animal participants.

The code used for the experiments as well as the raw experiment data results are available at \url{https://github.com/BenCrulis/binary_nn_extended}.
The datasets used in the experiments are available at \url{https://github.com/fastai/imagenette}, \cite{Krizhevsky2009} and \cite{Quattoni2010}.

\bibliography{references, additional-references, cyril}


\begin{thebibliography}{25}
\ifx \bisbn   \undefined \def \bisbn  #1{ISBN #1}\fi
\ifx \binits  \undefined \def \binits#1{#1}\fi
\ifx \bauthor  \undefined \def \bauthor#1{#1}\fi
\ifx \batitle  \undefined \def \batitle#1{#1}\fi
\ifx \bjtitle  \undefined \def \bjtitle#1{#1}\fi
\ifx \bvolume  \undefined \def \bvolume#1{\textbf{#1}}\fi
\ifx \byear  \undefined \def \byear#1{#1}\fi
\ifx \bissue  \undefined \def \bissue#1{#1}\fi
\ifx \bfpage  \undefined \def \bfpage#1{#1}\fi
\ifx \blpage  \undefined \def \blpage #1{#1}\fi
\ifx \burl  \undefined \def \burl#1{\textsf{#1}}\fi
\ifx \doiurl  \undefined \def \doiurl#1{\url{https://doi.org/#1}}\fi
\ifx \betal  \undefined \def \betal{\textit{et al.}}\fi
\ifx \binstitute  \undefined \def \binstitute#1{#1}\fi
\ifx \binstitutionaled  \undefined \def \binstitutionaled#1{#1}\fi
\ifx \bctitle  \undefined \def \bctitle#1{#1}\fi
\ifx \beditor  \undefined \def \beditor#1{#1}\fi
\ifx \bpublisher  \undefined \def \bpublisher#1{#1}\fi
\ifx \bbtitle  \undefined \def \bbtitle#1{#1}\fi
\ifx \bedition  \undefined \def \bedition#1{#1}\fi
\ifx \bseriesno  \undefined \def \bseriesno#1{#1}\fi
\ifx \blocation  \undefined \def \blocation#1{#1}\fi
\ifx \bsertitle  \undefined \def \bsertitle#1{#1}\fi
\ifx \bsnm \undefined \def \bsnm#1{#1}\fi
\ifx \bsuffix \undefined \def \bsuffix#1{#1}\fi
\ifx \bparticle \undefined \def \bparticle#1{#1}\fi
\ifx \barticle \undefined \def \barticle#1{#1}\fi
\bibcommenthead
\ifx \bconfdate \undefined \def \bconfdate #1{#1}\fi
\ifx \botherref \undefined \def \botherref #1{#1}\fi
\ifx \url \undefined \def \url#1{\textsf{#1}}\fi
\ifx \bchapter \undefined \def \bchapter#1{#1}\fi
\ifx \bbook \undefined \def \bbook#1{#1}\fi
\ifx \bcomment \undefined \def \bcomment#1{#1}\fi
\ifx \oauthor \undefined \def \oauthor#1{#1}\fi
\ifx \citeauthoryear \undefined \def \citeauthoryear#1{#1}\fi
\ifx \endbibitem  \undefined \def \endbibitem {}\fi
\ifx \bconflocation  \undefined \def \bconflocation#1{#1}\fi
\ifx \arxivurl  \undefined \def \arxivurl#1{\textsf{#1}}\fi
\csname PreBibitemsHook\endcsname

\bibitem[\protect\citeauthoryear{Crulis et~al.}{2023}]{crulis2023alternatives}
\begin{bchapter}
\bauthor{\bsnm{Crulis}, \binits{B.}},
\bauthor{\bsnm{Serres}, \binits{B.}},
\bauthor{\bsnm{Runz}, \binits{C.D.}},
\bauthor{\bsnm{Venturini}, \binits{G.}}:
\bctitle{{Are alternatives to backpropagation useful for training Binary Neural
  Networks? An experimental study in image classification}},
pp. \bfpage{1171}--\blpage{1178}
(\byear{2023})
\end{bchapter}
\endbibitem

\bibitem[\protect\citeauthoryear{He et~al.}{2021}]{He2021}
\begin{barticle}
\bauthor{\bsnm{He}, \binits{S.}},
\bauthor{\bsnm{Meng}, \binits{H.}},
\bauthor{\bsnm{Zhou}, \binits{Z.}},
\bauthor{\bsnm{Liu}, \binits{Y.}},
\bauthor{\bsnm{Huang}, \binits{K.}},
\bauthor{\bsnm{Chen}, \binits{G.}}:
\batitle{An efficient gpu-accelerated inference engine for binary neural
  network on mobile phones}.
\bjtitle{Journal of Systems Architecture}
\bvolume{117},
\bfpage{102156}
(\byear{2021})
\doiurl{10.1016/j.sysarc.2021.102156}
\end{barticle}
\endbibitem

\bibitem[\protect\citeauthoryear{Zhou et~al.}{2017}]{Zhou2017}
\begin{barticle}
\bauthor{\bsnm{Zhou}, \binits{Y.}},
\bauthor{\bsnm{Redkar}, \binits{S.}},
\bauthor{\bsnm{Huang}, \binits{X.}}:
\batitle{Deep learning binary neural network on an fpga}.
\bjtitle{Midwest Symposium on Circuits and Systems}
\bvolume{2017-Augus},
\bfpage{281}--\blpage{284}
(\byear{2017})
\doiurl{10.1109/MWSCAS.2017.8052915}
\end{barticle}
\endbibitem

\bibitem[\protect\citeauthoryear{Zhao et~al.}{2022}]{Zhao2022}
\begin{barticle}
\bauthor{\bsnm{Zhao}, \binits{J.}},
\bauthor{\bsnm{Xu}, \binits{S.}},
\bauthor{\bsnm{Wang}, \binits{R.}},
\bauthor{\bsnm{Zhang}, \binits{B.}},
\bauthor{\bsnm{Guo}, \binits{G.}},
\bauthor{\bsnm{Doermann}, \binits{D.}},
\bauthor{\bsnm{Sun}, \binits{D.}}:
\batitle{Data-adaptive binary neural networks for efficient object detection
  and recognition}.
\bjtitle{Pattern Recognition Letters}
\bvolume{153},
\bfpage{239}--\blpage{245}
(\byear{2022})
\doiurl{10.1016/j.patrec.2021.12.012}
\end{barticle}
\endbibitem

\bibitem[\protect\citeauthoryear{min Qian and Xiang}{2019}]{Qian2019}
\begin{barticle}
\bauthor{\bsnm{Qian}, \binits{Y.-m.}},
\bauthor{\bsnm{Xiang}, \binits{X.}}:
\batitle{Binary neural networks for speech recognition}.
\bjtitle{Frontiers of Information Technology and Electronic Engineering}
\bvolume{20},
\bfpage{701}--\blpage{715}
(\byear{2019})
\doiurl{10.1631/FITEE.1800469}
\end{barticle}
\endbibitem

\bibitem[\protect\citeauthoryear{Ojeda et~al.}{2020}]{ojeda2020}
\begin{bchapter}
\bauthor{\bsnm{Ojeda}, \binits{F.C.}},
\bauthor{\bsnm{Bisulco}, \binits{A.}},
\bauthor{\bsnm{Kepple}, \binits{D.}},
\bauthor{\bsnm{Isler}, \binits{V.}},
\bauthor{\bsnm{Lee}, \binits{D.D.}}:
\bctitle{On-device event filtering with binary neural networks for pedestrian
  detection using neuromorphic vision sensors},
pp. \bfpage{3084}--\blpage{3088}
(\byear{2020}).
\doiurl{10.1109/ICIP40778.2020.9191148}
\end{bchapter}
\endbibitem

\bibitem[\protect\citeauthoryear{Chen et~al.}{2020}]{Chen2020}
\begin{barticle}
\bauthor{\bsnm{Chen}, \binits{G.}},
\bauthor{\bsnm{Ling}, \binits{Y.}},
\bauthor{\bsnm{He}, \binits{T.}},
\bauthor{\bsnm{Meng}, \binits{H.}},
\bauthor{\bsnm{He}, \binits{S.}},
\bauthor{\bsnm{Zhang}, \binits{Y.}},
\bauthor{\bsnm{Huang}, \binits{K.}}:
\batitle{Stereoengine: An fpga-based accelerator for real-time high-quality
  stereo estimation with binary neural network}.
\bjtitle{IEEE Transactions on Computer-Aided Design of Integrated Circuits and
  Systems}
\bvolume{39},
\bfpage{4179}--\blpage{4190}
(\byear{2020})
\doiurl{10.1109/TCAD.2020.3012864}
\end{barticle}
\endbibitem

\bibitem[\protect\citeauthoryear{Daghero et~al.}{2021}]{Daghero2021}
\begin{botherref}
\oauthor{\bsnm{Daghero}, \binits{F.}},
\oauthor{\bsnm{Xie}, \binits{C.}},
\oauthor{\bsnm{Pagliari}, \binits{D.J.}},
\oauthor{\bsnm{Burrello}, \binits{A.}},
\oauthor{\bsnm{Castellano}, \binits{M.}},
\oauthor{\bsnm{Gandolfi}, \binits{L.}},
\oauthor{\bsnm{Calimera}, \binits{A.}},
\oauthor{\bsnm{MacIi}, \binits{E.}},
\oauthor{\bsnm{Poncino}, \binits{M.}}:
Ultra-compact binary neural networks for human activity recognition on risc-v
  processors.
Proceedings of the 18th ACM International Conference on Computing Frontiers
  2021, CF 2021,
3--11
(2021)
\doiurl{10.1145/3457388.3458656}
\end{botherref}
\endbibitem

\bibitem[\protect\citeauthoryear{Fasfous et~al.}{2021}]{Fasfous2021}
\begin{bchapter}
\bauthor{\bsnm{Fasfous}, \binits{N.}},
\bauthor{\bsnm{Vemparala}, \binits{M.R.}},
\bauthor{\bsnm{Frickenstein}, \binits{A.}},
\bauthor{\bsnm{Frickenstein}, \binits{L.}},
\bauthor{\bsnm{Badawy}, \binits{M.}},
\bauthor{\bsnm{Stechele}, \binits{W.}}:
\bctitle{Binarycop: Binary neural network-based covid-19 face-mask wear and
  positioning predictor on edge devices},
pp. \bfpage{108}--\blpage{115}
(\byear{2021}).
\doiurl{10.1109/IPDPSW52791.2021.00024}
\end{bchapter}
\endbibitem

\bibitem[\protect\citeauthoryear{Hubara et~al.}{2016}]{Hubara2016a}
\begin{bchapter}
\bauthor{\bsnm{Hubara}, \binits{I.}},
\bauthor{\bsnm{Courbariaux}, \binits{M.}},
\bauthor{\bsnm{Soudry}, \binits{D.}},
\bauthor{\bsnm{El-Yaniv}, \binits{R.}},
\bauthor{\bsnm{Bengio}, \binits{Y.}}:
\bctitle{Binarized neural networks}.
In: \bbtitle{Advances in Neural Information Processing Systems},
pp. \bfpage{4114}--\blpage{4122}
(\byear{2016})
\end{bchapter}
\endbibitem

\bibitem[\protect\citeauthoryear{Yuan and Agaian}{2023}]{yuan2023BNNSurvey}
\begin{botherref}
\oauthor{\bsnm{Yuan}, \binits{C.}},
\oauthor{\bsnm{Agaian}, \binits{S.S.}}:
A comprehensive review of binary neural network.
Artificial Intelligence Review,
1--65
(2023)
\end{botherref}
\endbibitem

\bibitem[\protect\citeauthoryear{Blum and Rivest}{1992}]{BLUM1992117}
\begin{barticle}
\bauthor{\bsnm{Blum}, \binits{A.L.}},
\bauthor{\bsnm{Rivest}, \binits{R.L.}}:
\batitle{Training a 3-node neural network is np-complete}.
\bjtitle{Neural Networks}
\bvolume{5}(\bissue{1}),
\bfpage{117}--\blpage{127}
(\byear{1992})
\doiurl{10.1016/S0893-6080(05)80010-3}
\end{barticle}
\endbibitem

\bibitem[\protect\citeauthoryear{Lillicrap et~al.}{2016}]{Lillicrap2016}
\begin{barticle}
\bauthor{\bsnm{Lillicrap}, \binits{T.P.}},
\bauthor{\bsnm{Cownden}, \binits{D.}},
\bauthor{\bsnm{Tweed}, \binits{D.B.}},
\bauthor{\bsnm{Akerman}, \binits{C.J.}}:
\batitle{Random synaptic feedback weights support error backpropagation for
  deep learning}.
\bjtitle{Nature Communications}
\bvolume{7},
\bfpage{1}--\blpage{10}
(\byear{2016})
\doiurl{10.1038/ncomms13276}
\end{barticle}
\endbibitem

\bibitem[\protect\citeauthoryear{Nøkland}{2016}]{Nokland2016}
\begin{bchapter}
\bauthor{\bsnm{Nøkland}, \binits{A.}}:
\bctitle{Direct feedback alignment provides learning in deep neural networks},
pp. \bfpage{1045}--\blpage{1053}
(\byear{2016})
\end{bchapter}
\endbibitem

\bibitem[\protect\citeauthoryear{Frenkel et~al.}{2021}]{Frenkel2021}
\begin{barticle}
\bauthor{\bsnm{Frenkel}, \binits{C.}},
\bauthor{\bsnm{Lefebvre}, \binits{M.}},
\bauthor{\bsnm{Bol}, \binits{D.}}:
\batitle{Learning without feedback: Fixed random learning signals allow for
  feedforward training of deep neural networks}.
\bjtitle{Frontiers in Neuroscience}
\bvolume{15},
\bfpage{1}--\blpage{13}
(\byear{2021})
\doiurl{10.3389/fnins.2021.629892}
\end{barticle}
\endbibitem

\bibitem[\protect\citeauthoryear{Ma et~al.}{2020}]{KurtMa2020}
\begin{bchapter}
\bauthor{\bsnm{Ma}, \binits{W.D.K.}},
\bauthor{\bsnm{Lewis}, \binits{J.P.}},
\bauthor{\bsnm{Kleijn}, \binits{W.B.}}:
\bctitle{The hsic bottleneck: Deep learning without back-propagation},
pp. \bfpage{5085}--\blpage{5092}
(\byear{2020}).
\doiurl{10.1609/aaai.v34i04.5950}
\end{bchapter}
\endbibitem

\bibitem[\protect\citeauthoryear{Pogodin and Latham}{2020}]{Pogodin2020}
\begin{bchapter}
\bauthor{\bsnm{Pogodin}, \binits{R.}},
\bauthor{\bsnm{Latham}, \binits{P.E.}}:
\bctitle{Kernelized information bottleneck leads to biologically plausible
  3-factor hebbian learning in deep networks},
vol. \bseriesno{2020-Decem}
(\byear{2020})
\end{bchapter}
\endbibitem

\bibitem[\protect\citeauthoryear{Kohan et~al.}{2023}]{Kohan2022}
\begin{botherref}
\oauthor{\bsnm{Kohan}, \binits{A.}},
\oauthor{\bsnm{Rietman}, \binits{E.A.}},
\oauthor{\bsnm{Siegelmann}, \binits{H.T.}}:
Signal propagation: The framework for learning and inference in a forward pass.
IEEE Transactions on Neural Networks and Learning Systems,
1--12
(2023)
\doiurl{10.1109/TNNLS.2022.3230914}
\end{botherref}
\endbibitem

\bibitem[\protect\citeauthoryear{Lansdell et~al.}{2020}]{Lansdell2020}
\begin{bchapter}
\bauthor{\bsnm{Lansdell}, \binits{B.J.}},
\bauthor{\bsnm{Prakash}, \binits{P.R.}},
\bauthor{\bsnm{Körding}, \binits{K.P.}}:
\bctitle{Learning to solve the credit assignment problem}.
\bpublisher{OpenReview.net}, \blocation{???}
(\byear{2020}).
\burl{https://openreview.net/forum?id=ByeUBANtvB}
\end{bchapter}
\endbibitem

\bibitem[\protect\citeauthoryear{Akshat and Prasad}{2022}]{Akshat2022}
\begin{barticle}
\bauthor{\bsnm{Akshat}, \binits{G.}},
\bauthor{\bsnm{Prasad}, \binits{N.R.}}:
\batitle{Blind descent: A prequel to gradient descent}.
\bjtitle{Lecture Notes in Electrical Engineering}
\bvolume{783},
\bfpage{473}--\blpage{479}
(\byear{2022})
\doiurl{10.1007/978-981-16-3690-5_41}
\end{barticle}
\endbibitem

\bibitem[\protect\citeauthoryear{Simonyan and Zisserman}{2015}]{Simonyan2015}
\begin{botherref}
\oauthor{\bsnm{Simonyan}, \binits{K.}},
\oauthor{\bsnm{Zisserman}, \binits{A.}}:
Very deep convolutional networks for large-scale image recognition.
3rd International Conference on Learning Representations, ICLR 2015 -
  Conference Track Proceedings,
1--14
(2015)
\end{botherref}
\endbibitem

\bibitem[\protect\citeauthoryear{Sandler et~al.}{2018}]{Sandler2018}
\begin{botherref}
\oauthor{\bsnm{Sandler}, \binits{M.}},
\oauthor{\bsnm{Howard}, \binits{A.}},
\oauthor{\bsnm{Zhu}, \binits{M.}},
\oauthor{\bsnm{Zhmoginov}, \binits{A.}},
\oauthor{\bsnm{Chen}, \binits{L.C.}}:
Mobilenetv2: Inverted residuals and linear bottlenecks.
Proceedings of the IEEE Computer Society Conference on Computer Vision and
  Pattern Recognition,
4510--4520
(2018)
\doiurl{10.1109/CVPR.2018.00474}
\end{botherref}
\endbibitem

\bibitem[\protect\citeauthoryear{Tolstikhin et~al.}{2021}]{Tolstikhin2021}
\begin{barticle}
\bauthor{\bsnm{Tolstikhin}, \binits{I.}},
\bauthor{\bsnm{Houlsby}, \binits{N.}},
\bauthor{\bsnm{Kolesnikov}, \binits{A.}},
\bauthor{\bsnm{Beyer}, \binits{L.}},
\bauthor{\bsnm{Zhai}, \binits{X.}},
\bauthor{\bsnm{Unterthiner}, \binits{T.}},
\bauthor{\bsnm{Yung}, \binits{J.}},
\bauthor{\bsnm{Steiner}, \binits{A.}},
\bauthor{\bsnm{Keysers}, \binits{D.}},
\bauthor{\bsnm{Uszkoreit}, \binits{J.}},
\bauthor{\bsnm{Lucic}, \binits{M.}},
\bauthor{\bsnm{Dosovitskiy}, \binits{A.}}:
\batitle{Mlp-mixer: An all-mlp architecture for vision}.
\bjtitle{Advances in Neural Information Processing Systems}
\bvolume{29},
\bfpage{24261}--\blpage{24272}
(\byear{2021})
\end{barticle}
\endbibitem

\bibitem[\protect\citeauthoryear{Krizhevsky}{2009}]{Krizhevsky2009}
\begin{botherref}
\oauthor{\bsnm{Krizhevsky}, \binits{A.}}:
Learning Multiple Layers of Features from Tiny Images
(2009)
\end{botherref}
\endbibitem

\bibitem[\protect\citeauthoryear{Quattoni and Torralba}{2010}]{Quattoni2010}
\begin{botherref}
\oauthor{\bsnm{Quattoni}, \binits{A.}},
\oauthor{\bsnm{Torralba}, \binits{A.}}:
Recognizing indoor scenes.
IEEE Conference on Computer Vision and Pattern Recognition (CVPR),
413--420
(2010)
\doiurl{10.1109/cvpr.2009.5206537}
\end{botherref}
\endbibitem

\end{thebibliography}

\end{document}